\pdfoutput=1
\documentclass[11pt]{article}

\usepackage[final]{acl}

\usepackage{times}
\usepackage{latexsym}
 \usepackage{amsmath}
\usepackage{booktabs}
\usepackage{url}

\usepackage{enumitem}
\setlist[enumerate]{noitemsep}
\usepackage{subfigure}

\usepackage[T1]{fontenc}
\usepackage[utf8]{inputenc}

\usepackage{microtype}

\usepackage{inconsolata}

\usepackage{graphicx}

\title{The Behavior Gap: Evaluating Zero-shot LLM Agents\\in Complex Task-Oriented Dialogs}

\author{Avinash Baidya, Kamalika Das, Xiang Gao \\
        Intuit AI Research \\
        2700 Coast Avenue, Mountain View, CA 94043 \\
        \texttt{\{avinash\_baidya, kamalika\_das, xiang\_gao\}@intuit.com}}

\begin{document}
\maketitle

\begin{abstract}

Large Language Model (LLM)-based agents have significantly impacted Task-Oriented Dialog Systems (TODS) but continue to face notable performance challenges, especially in zero-shot scenarios. While prior work has noted this performance gap, the behavioral factors driving the performance gap remain under-explored.
This study proposes a comprehensive evaluation framework to quantify the behavior gap between AI agents and human experts, focusing on discrepancies in dialog acts, tool usage, and knowledge utilization.  
Our findings reveal that this behavior gap is a critical factor negatively impacting the performance of LLM agents. 
Notably, as task complexity increases, the behavior gap widens (correlation: 0.963), leading to a degradation of agent performance on complex task-oriented dialogs. 
For the most complex task in our study, even the GPT-4o-based agent exhibits low alignment with human behavior, with low F1 scores for dialog acts (0.464), excessive and often misaligned tool usage with a F1 score of 0.139, and ineffective usage of external knowledge. 
Reducing such behavior gaps leads to significant performance improvement (24.3\% on average).
This study highlights the importance of comprehensive behavioral evaluations and improved alignment strategies to enhance the effectiveness of LLM-based TODS in handling complex tasks. \footnote{The code will be released at \url{https://github.com/intuit-ai-research/behavior-gap}.}

\end{abstract}

\section{Introduction}

In recent years, advancements in artificial intelligence, particularly in Large Language Models (LLMs) \citep{brown2020gpt3, ouyang2022training, achiam2023gpt4, touvron2023llama, team2023gemini, bai2023qwen, jiang2023mistral, elizabeth2024large, guo2025deepseek}, and LLM-based agents \citep{wang2024survey, xi2025rise, yaoreact}, have garnered significant attention for Task-Oriented Dialog Systems (TODS).
While LLM agents have the promise of eliminating complex modular designs in traditional TODS~\citep{xu2024rethinking},
recent studies, however, have shown that LLM agents, particularly in zero-shot scenarios, struggle to perform optimally \citep{elizabeth2024large, heck2023chatgpt, zhang2023sgp, hudecek_are_2023}.

Though several studies have identified this performance gap, there is limited work investigating the behavioral causes. By considering three specific dialog acts, \citet{shaikh2024grounding} found that LLMs generate language with significantly less grounding compared to humans. However, this is not necessarily the only source of the \textbf{behavior gap}—the discrepancies between LLM agents and human experts. A comprehensive study of the behavior gap is necessary. This involves not only identifying the dimensions along which these discrepancies occur but also understanding the factors that may affect this gap, such as task complexity and the choice of LLMs. Additionally, it is important to quantify the impact of these discrepancies on agent performance.

In this study, we propose a comprehensive evaluation framework to measure the behavior gap across three key behavioral dimensions. One dimension is \textbf{dialog acts}, traditionally used to represent an utterance at the level of illocutionary force \cite{austin1975things, stolcke2000dialogue, mezza2018iso}. The second dimension is \textbf{tool usage}, and the third is \textbf{external knowledge usage}, both included due to their crucial role in modern LLM agents \cite{qin2024tool, lewis2020retrieval, zhang2022retgen}. Our framework operates in a teacher-forcing setting \citep{williams1989learning} to avoid dependence on user simulators, which may introduce additional discrepancies, and could be implementation-specific.

Our analysis reveals significant behavioral gaps between LLM agents and human experts across three tasks of varying complexity levels. Specifically, LLMs exhibit misalignment in dialog acts, excessive but often incorrect tool usage, and inefficient representation of external knowledge. These gaps become especially pronounced as we move from relatively simple slot-filling tasks to more complex ones, particularly in the Product Customer Support (PCS) dataset. PCS is a new TOD dataset studied in this work that captures real-world customer-human expert interactions. Furthermore, our analysis reveals that the behavior gap negatively impacts LLM agent performance, and reducing this gap leads to statistically significant improvements.

In summary, our contributions are as follows:
\begin{itemize}
    \item We propose an evaluation framework to comprehensively quantify the behavior gap between human experts and LLM agents. This framework identifies discrepancies across three key behavioral dimensions: dialog acts, tool usage, and external knowledge usage.
    \item Using this framework, we uncover key insights regarding LLM agent behavior: (a) LLM agents differ significantly from humans in their choice of dialog acts and tool usage across models and tasks; (b) these gaps widen as task complexity increases or as the model size decreases; and (c) LLM agents tend to copy external knowledge verbatim rather than synthesizing it as humans do.
    \item We demonstrate that the behavior gap is both statistically correlated with task performance and a critical factor limiting performance—reducing this gap directly improves LLM agent performance, particularly in complex tasks.
\end{itemize}

\section{Related Work}

\paragraph{LLM Agents for TODS} 

\citet{xu2024rethinking} introduced AutoTOD, a zero-shot autonomous TOD agent that eliminates the complex modular components of traditional systems, relying solely on instruction-following LLMs. \citet{elizabeth2024large} introduced a ReAct-based LLM agent for TODS. \citet{dong2025protod} proposed ProTOD, which incorporates a passive-to-proactive policy planner to enhance system capabilities. \citet{zhang2023sgp} emphasized the integration of domain-specific knowledge, such as schemas, to enhance system performance. Additionally, \citet{gupta2024dard} presented DARD, a multi-agent framework that deploys domain-specific agents and LLMs of varying sizes to handle diverse tasks effectively. While these agent-centric works demonstrate promising ways to eliminate traditional TODS pipelines, our study complements them by showing that such architectures still exhibit a measurable "behavior gap" from human experts. We introduce a framework for quantifying that gap across dialog acts, tool use, and knowledge usage.

\paragraph{Behavior and Performance Evaluation}

FED \citep{mehri2020fed} provides an evaluation framework for assessing fine-grained dialog quality using DialoGPT \citep{zhang2020dialogpt}. However, since DialoGPT is trained on Reddit data, its applicability to TODS is limited. Recent advancements have explored LLM-as-a-judge approaches \citep{zheng2023judging, liu2023g} as alternative evaluation methods for dialog and other complex language generation tasks that are traditionally difficult to assess. Despite these efforts, few studies have focused on the behavioral analysis of LLM agents. One notable study \cite{shaikh2024grounding} found that LLMs exhibit less conversational grounding compared to humans. However, conversational grounding is only one of several factors contributing to the behavior gap. We build on these works by introducing a comprehensive behavior-centric analysis that spans dialog acts, tool selection, and knowledge grounding, providing deeper insight into how specific behavioral gaps influence overall TOD agent performance.

\section{Method}

In this work, we introduce a comprehensive framework designed to systematically analyze the behavior gap between LLM-based agents and human experts. This framework includes: 
\begin{enumerate}
    \item Metrics to quantify the behavior gap across dialog acts, tool usage, and knowledge integration (Section~\ref{sec:method-behavior}).
    \item Teacher-forcing approach for controlled evaluation (Section~\ref{sec:method-performance}).
    \item Performance measures that reveal the relationship between the behavior gap and agent performance (Section~\ref{sec:method-performance}).
    \item Task complexity measurement with two complementary metrics (Section~\ref{sec:task_complexity}).
\end{enumerate}

Using this framework, we evaluated the behavior and performance of LLM agents (Section~\ref{sec:method-agent}) across three tasks of varying complexity (Section~\ref{sec:method-datasets}).

\subsection{Datasets}
\label{sec:method-datasets}

To study the relationship between task complexity, behavior, and task performance, we selected three TOD datasets: \textbf{MultiWOZ} \cite{budzianowski2018multiwoz}, \textbf{SpokenWOZ} \cite{si2024spokenwoz}, and \textbf{PCS}, a product customer support dataset. 

The MultiWOZ dataset~\cite{budzianowski2018multiwoz} is a widely used benchmark for evaluating TOD systems, consisting of human-human written dialogs in multiple domains. SpokenWOZ~\cite{si2024spokenwoz} builds upon MultiWOZ by introducing human-to-human spoken conversations.

The Product Customer Support (PCS) dataset is a private dataset and represents the most complex scenario in our study. It consists of real-world transcribed spoken conversations between customers and human expert agents in a customer support setting. These dialogs involve tasks that require multi-step reasoning, such as troubleshooting, and display a variety of intents and actions from both the agent and user. A snippet from a sample conversation is demonstrated in Table~\ref{tab:pcs_snippet}.

These datasets were chosen to represent an increasing amount of complexity, defined by factors such as the average number of dialog turns and the variety of intents and actions (see Section~\ref{sec:task_complexity} and Fig.~\ref{fig:task_complexity} for a quantitative measure of their complexity). This progression in complexity allows us to systematically evaluate how behavior gap scales with task difficulty and how it impacts the overall task performance. Key statistics of these datasets are presented in Table~\ref{tab:dataset}. See \hyperref[sec:appendix]{Appendix} for additional details.

\begin{table}
    \centering
    \small
    \begin{tabular}{lccc}
        \toprule
         & MultiWOZ & SpokenWOZ & PCS \\
          \hline
        Chats & 1000 & 987 & 53 \\ 
        Turns/Chat & 14.7 & 35.6 & 120.2 \\
        Words/Turn & 13.4 & 11.0 & 11.8 \\
        Valid Slots & 24 & 36 & $\infty$ \\
        Valid Tools & 8 & 9 & 4 \\
        \bottomrule
    \end{tabular}
    \caption{Key statistics of the test datasets} 
    \label{tab:dataset}
\end{table}

\subsection{LLM Agent Design and Implementation}
\label{sec:method-agent}

We employed a \textbf{Zero-shot agent} with access to external tools. 
The Zero-shot agent operates in a multi-turn dialog setting, where it processes user inputs, reasons about the task requirements, and invokes appropriate tools to achieve task objectives. The agent does not rely on task-specific fine-tuning. It leverages pre-trained language models for natural language understanding and reasoning.  In this work, we employed three state-of-the-art pretrained language model: \textbf{GPT-4o}~\cite{hurst2024gpt} (\texttt{gpt-4o-2024-05-13}), \textbf{GPT-3.5 Turbo}~(\texttt{gpt-35-turbo-0125}), and \textbf{LLaMA-3.3-70B-Instruct}~\cite{dubey2024llama}\footnote{License: \url{https://github.com/meta-llama/llama3}}.

\subsubsection{Tools}

For each dataset, we tailored the agent's tool set and system prompt to align with the specific requirements of the task. Below, we describe the available tools for each task (see \hyperref[sec:appendix]{Appendix} for details).

\paragraph{MultiWOZ.}
For the MultiWOZ task, the agent was equipped with tools to allow information retrieval from the MultiWOZ database~\cite{budzianowski2018multiwoz} and booking across five domains: \texttt{BookHotel}, \texttt{BookTaxi}, \texttt{BookRestaurant}, \texttt{BookTrain},  \texttt{FindHotels}, \texttt{FindRestaurants}, \texttt{FindAttractions}, and \texttt{FindTrains}.

\paragraph{SpokenWOZ.}
The SpokenWOZ task builds upon MultiWOZ, incorporating all the MultiWOZ tools along with an additional \texttt{BookParking} tool for the agent.

\paragraph{PCS.} 
For this dataset, the agent was equipped with four specialized tools, based on human tool usage patterns, that allowed the agent to look up external knowledge base  and simulate screen share, customer information look up, and escalation or transfer to other departments: \texttt{ScreenShare}, \texttt{KnowledgeLookup},  \texttt{CustomerInfoLookup}, and \texttt{EscalateOrTransfer}.

\subsubsection{Planning and Reasoning}

Inspired from \citet{elizabeth2024large, zhao2024expel}, we employed the ReAct framework~\cite{yaoreact}, which combines Chain of Thought style reasoning and acting capabilities.
This framework is designed to handle complex task-oriented dialogs by utilizing natural language understanding, reasoning, and action execution. The agent was implemented using LangGraph~\cite{langgraph2024}.

\subsection{Behavior Evaluation }
\label{sec:method-behavior}

To systematically analyze and compare the behavior of LLM agents and human experts, we conducted a detailed turn-by-turn analysis of their responses. The analysis focused on three key aspects: the dialog acts present in each response, the tools invoked during the conversation, and the use of external knowledge to address user queries. 
These are among the basic elements that affect the multi-turn dialog strategy.

\subsubsection{Dialog Acts}
\label{sec:dialog_acts}

We employed two well-known frameworks~\cite{budzianowski2018multiwoz, mezza2018iso} to annotate the turn-by-turn dialog acts present in each response:
\begin{itemize}
    \item \textbf{Task-specific WOZ Framework}: This dialog act taxonomy has been widely used to analyze dialog acts in tasks such as MultiWOZ, SpokenWOZ, and similar TOD datasets~\cite{budzianowski2018multiwoz, si2024spokenwoz, rastogi_towards_2020}.. Hereafter, we will refer to this framework as the "WOZ" framework. It comprehensively captures all the key conversational scenarios across multiple domains present in these tasks. Specifically, we considered 10 dialog act types introduced in \citealt{budzianowski2018multiwoz}: \texttt{inform, request, select, recommend, nooffer, offerbook, book, nobook, greet}, and \texttt{reqmore} (see \hyperref[sec:appendix]{Appendix} for details).

    \item \textbf{Task-independent ISO framework}: This framework provides a task-independent and comprehensive dialog act taxonomy~\cite{mezza2018iso, bunt2012iso} grounded in the Dynamic Interpretation Theory (DIT) for dialogs~\cite{bunt2012dynamic}. Following \citet{mezza2018iso}, the dialog act types included in this framework were: \texttt{set\_q, prop\_q, choice\_q, other\_q, inform, commissives, directives, salutation, apology, thanking}, and \texttt{feedback}. 
    % (see \hyperref[sec:appendix]{Appendix} for details)
    These acts were designed to capture all key conversational scenarios for open-domain human-machine conversations~\cite{mezza2018iso, bunt2012iso}.
\end{itemize}

These two frameworks combined allowed us to systematically evaluate the dialog acts for both LLM agents and human experts across tasks. We used the WOZ framework to evaluate the MultiWOZ and SpokenWOZ tasks, since this framework was designed to cover all possible scenarios in these two tasks. However, we adopted the more flexible ISO framework for the open-ended PCS task.

\paragraph{LLM-Based Dialog Act Classifiers.}
To analyze the responses based on the above frameworks, we employed two GPT-4o-based few-shot classifiers for the two (WOZ and ISO) frameworks, respectively. Given the user input and the corresponding system response, the output of the classifiers consisted of a list of predicted dialog act types present in the system response. To annotate an entire dialog, each set of turns (the user input and the corresponding system response) were processed one at a time. See \hyperref[sec:appendix]{Appendix} for details on the classifiers.

Both classifiers were validated against ground-truth annotations from the MultiWOZ~\cite{budzianowski_multiwoz_2020} dataset for the WOZ framework and the DialogBank~\cite{bunt2019dialogbank} dataset for the ISO framework. Dialog act classification being a multi-label classification task where each turn can carry multiple labels (dialog act types), was evaluated using the micro-F1 score. This metric was chosen because it aggregates true positives, false positives, and false negatives across all labels—offering a robust performance measure that mitigates class imbalance. The WOZ and ISO classifiers achieved overall micro-F1 score of 0.771 and 0.745, respectively. Both classifiers performed significantly better than chance\footnote{A random classifier achieved a micro-F1 score of 0.27 and 0.09 for the WOZ and ISO dialog act validation, respectively, which is substantially lower than our classifiers' scores of 0.771 and 0.745.}, demonstrating their reliability.

\subsubsection{Tool Usage}
\label{sec:tool_use}
To compare tool usage patterns between the LLM agent and the human expert, we developed a GPT-4o-based few-shot tool classifier. The classifier was designed to annotate tools used in human expert responses, as ground-truth tool usage for LLM agents was already available. Since each task involved a unique set of tools, we created task-specific classifiers tailored for each of the three tasks: MultiWOZ, SpokenWOZ, and PCS. Additional details are provided in the \hyperref[sec:appendix]{Appendix}.

The ground-truth tool usage from LLM agent responses was used to validate the three classifiers. Like dialog act classification, this task is a multi-label classification problem, and we used the micro-F1 score for evaluation. The MultiWOZ, SpokenWOZ, and PCS classifiers achieved micro-F1 scores of 0.893, 0.898, and 0.748, respectively. All classifiers performed significantly above chance\footnote{A random classifier achieved a micro-F1 score of 0.19, 0.13, and 0.33 for the MultiWOZ, SpokenWOZ, and PCS tool validation, respectively, which is substantially lower than our classifiers' scores of 0.893, 0.898, and 0.748.}, confirming their reliability.

\subsubsection{External Knowledge Usage}
\label{sec:knowledge_use}

To compare how LLM agents and human experts utilize retrieved external knowledge, we focused on turns where both agents and humans used tools designed to specifically access external knowledge: the search tools for querying the MultiWOZ database in the MultiWOZ and SpokenWOZ tasks, and the knowledge lookup tool for the PCS task. We employed two metrics to analyze the retrieved knowledge usage:

\begin{enumerate}
    \item \textbf{ROUGE-1 Precision}: Measures how much of the generated response was directly copied from the retrieved knowledge.
    \item \textbf{Compression Ratio}: Measures the efficiency of condensing retrieved knowledge into the response, defined as:
    \begin{equation*}
        \text{Compression ratio} = 1 - \frac{\text{Response length}}{\text{Knowledge length}}.
    \end{equation*}
    where length is measured by the number of words. We assumed that human experts retrieved the same external knowledge as LLM agents, so the knowledge length was identical for both \footnote{
    For MultiWOZ and SpokenWOZ tasks, this assumption is reasonable as the external knowledge is strictly limited to the MultiWOZ database. 
    For the PCS task, it remains valid since both human experts and LLM agents have access to the same \texttt{KnowledgeLookup} tool.
    }.
\end{enumerate}

\subsection{Performance Analysis}
\label{sec:method-performance}

To systematically evaluate the effectiveness of LLM agents in TODS, we developed a framework comprising two key components: a teacher-forcing evaluation approach to compare LLM and human behavior in identical contexts, and an LLM-based evaluator to assess response quality across multiple dimensions.

\paragraph{The Teacher-Forcing Approach.} Existing TODS studies often employ a user simulator \cite{li2016user, li2017end, guo2018dialog} to evaluate the final performance of an LLM agent. User simulators, while useful, may fail to capture the nuanced nature of human behavior. This failure can cause errors that can compound across multiple turns, potentially amplifying errors in performance evaluation. 
To address this, we adopt a teacher-forcing~\cite{williams1989learning} evaluation approach. Originally referring to feeding ground-truth outputs back into a model during training, we adapt this concept by providing the previous conversation between the human user and human expert as context. 
Assuming the list $\{a_0, u_0, a_1, u_1, \ldots, a_n, u_n\}$ represents the utterances between the human expert ($a_i$) and human user ($u_i$) in time order, we feed $\{a_0, u_0, a_1, u_1, \ldots, u_{t-1}\}$ as the context to the LLM agent to obtain the generated response $g_t$, to be compared with the human agent response $a_t$. 
This allows us to directly compare the LLM agent’s behavior with that of the human expert in the same context.

\paragraph{Performance Evaluator.} We implemented an evaluator using GPT-4o to assess the agent performance at the turn level. The evaluator received the user input, the conversation history up to the given turn, and the corresponding agent’s response. It rated the agent's response across four key aspects: dialog coherence (Coherence; \citet{venkatesh2018evaluating}), information detail and precision (Specificity; \citet{adiwardana2020towards}), understanding user needs (Effectiveness; \citet{braggaar_evaluating_2024}) and user satisfaction (Satisfaction; \citet{feng2023schema}). Each metric was scored on a scale of 1 to 5, where 1 indicated a completely inadequate or incorrect response, and 5 represented a high-quality response meeting all criteria.

In the absence of ground-truth turn-level performance scores, we validated the evaluator by comparing aggregated dialog-level scores for each metric against ground-truth dialog success rates~\cite{nekvinda_shades_2021} on the MultiWOZ task. Dialogs with a success rate of 1 scored significantly higher (p < 0.05) across all metrics compared to those with a success rate of 0, demonstrating that the LLM-based evaluator is relatively reliable at assessing agent performance (see \hyperref[sec:appendix]{Appendix} for details).

\subsection{Task Complexity}
\label{sec:task_complexity}

\begin{figure}
    \centering
    \includegraphics[width=0.85\linewidth]{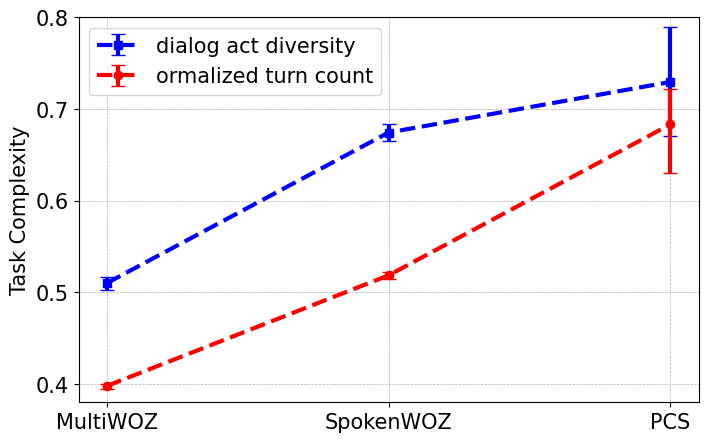}
    \caption{\textbf{Task complexity}. Comparison of task complexity across the three tasks measured by dialog act diversity and normalized turn count. Higher values indicate greater task complexity. Error bars denote 95\% confidence intervals.}
    \label{fig:task_complexity}
\end{figure}

We propose two complementary metrics to quantify task complexity for each dataset:

\begin{enumerate}
    \item \textbf{Normalized Turn Count}: Measures complexity in terms of dialog length: 
    \begin{equation*}
        \text{Normalized Turn Count} = \frac{\ln(1+t)}{\ln(1+t+C)},
    \end{equation*}
    where $t$ is average number of turns per chat and $C$ is a constant number\footnote{We set $C=1000$.}.
    
    \item \textbf{Dialog Act Diversity}: Measures complexity from the perspective of dialog acts:
    \begin{equation*}
        \text{Dialog Act Diversity} = d / d_\text{max},
    \end{equation*}
    where $d$ is the average number of distinct dialog acts per chat, and $d_\text{max}$ is the maximum number of available dialog acts\footnote{In our setting, $d_\text{max}=11$}. We use the ISO dialog act framework to ensure this metric is applicable across all tasks.
\end{enumerate}

As shown in Fig.~\ref{fig:task_complexity}, task complexity increases progressively from the slot-filling tasks, MultiWOZ and SpokenWOZ, to the more challenging PCS dataset. 

The two metrics used to assess task complexity are complementary. For instance, a lengthy conversation consisting solely of simple yes/no exchanges will result in a high \texttt{Normalized Turn Count} but remains fundamentally simple. In such cases, \texttt{Dialog Act Diversity} would remain low. Conversely,~\texttt{Dialog Act Diversity} can be high even for single-turn conversations if they involve complex responses featuring multiple dialog acts. In these scenarios, \texttt{Normalized Turn Count} provides additional context. 

Both metrics are normalized to a range between 0 and 1. To derive an overall measure of task complexity, we simply compute the average of the two metrics.

\section{Results}

\subsection{The Behavior Gap}

We analyzed the behavior gap between LLM agents and human experts across tasks of increasing complexity, focusing on differences in dialog acts, tool usage, and knowledge utilization patterns.

\begin{figure}[t]
    \centering
    \subfigure[]{
        \includegraphics[width=0.8\columnwidth]{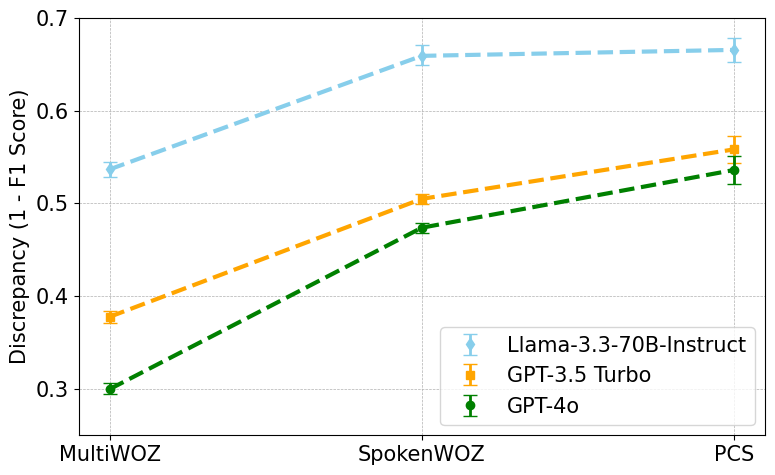}
    }
    \subfigure[]{
        \includegraphics[width=0.8\columnwidth]{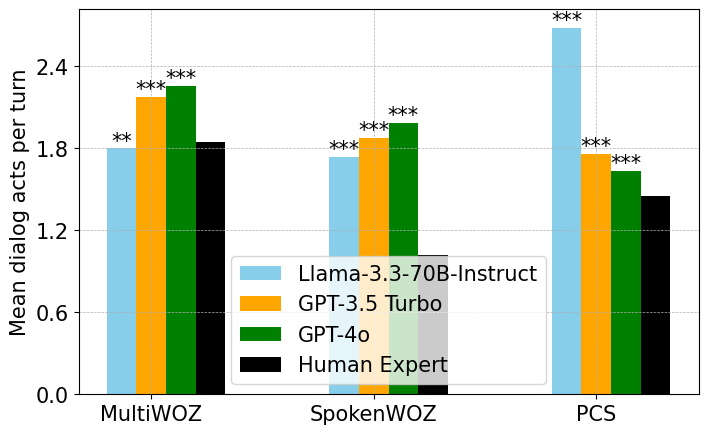}
    }
    \caption{
    \textbf{Dialog act gap}. Comparing the dialog act selection pattern across the three tasks between the human expert and the three LLM agents. \textbf{(a)} Overall discrepancy, measured as $1 - \text{micro-F1 score}$, between LLM agents and human experts. Error bars denote 95\% confidence intervals. \textbf{(b)} Mean number of dialog acts chosen per turn by agents and human experts. Asterisks denote statistical significance in the differences between agents and human experts.\protect\footnotemark % (black). 
    }
    \label{fig:dialog_gap}
\end{figure}
\footnotetext{Single asterisk (*) denotes $p < 0.05$, double asterisks (**) denote $p < 0.01$, and triple asterisks (***) denote $p < 0.001$, and \textit{n.s.} denotes a non-significant result.}
\begin{figure}[t]
    \centering
    \subfigure[]{
        \includegraphics[width=0.8\columnwidth]{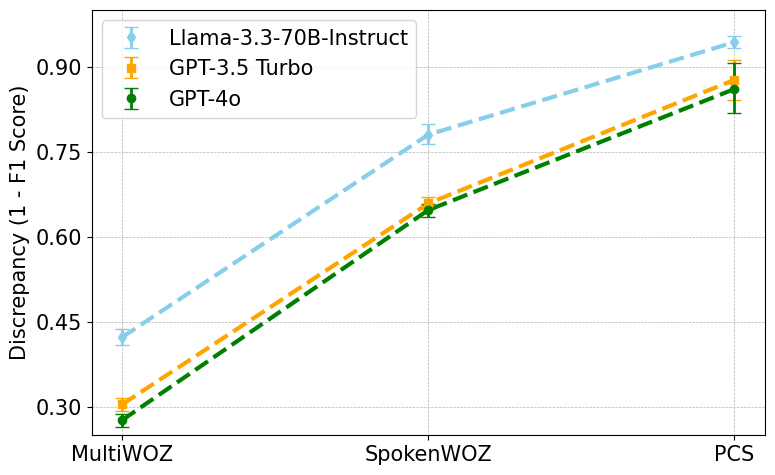}
    }
    \subfigure[]{
        \includegraphics[width=0.8\columnwidth]{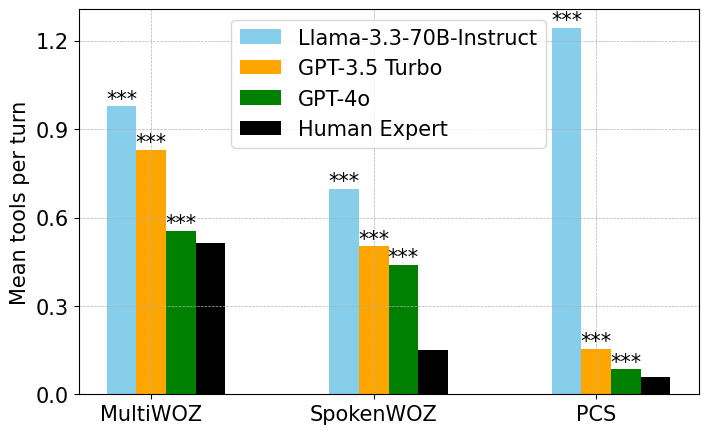}
    }
    \caption{
    \textbf{Tool usage gap}. Comparing the tool usage pattern across the three tasks between the human expert and the three LLM agents. \textbf{(a)} Overall discrepancy, computed as  $1 - \text{micro-F1 score}$, between LLM agents and human experts. Error bars denote 95\% confidence intervals. \textbf{(b)} Mean number of tools invoked per turn by LLM agents and human experts. Asterisks denote statistical significance in the differences between agents and human experts.
    }
    \label{fig:tool_gap}
\end{figure}

\subsubsection{Dialog Act Gap}

To compare the dialog act selection patterns of human experts and LLM agents, we annotated their turn-by-turn responses using the dialog act classifiers detailed in Section~\ref{sec:dialog_acts}. To evaluate the alignment of dialog acts between the human experts and LLM agents, we employed the micro-F1 score\footnote{This is treated as a multi-label classification problem with the dialog acts of the human experts considered as the target. The micro-F1 score reflects the degree to which the AI agent's responses match the human expert's targets.}. Discrepancy was quantified as $1 - \text{micro-F1 score}$. 

We observed that the overall discrepancy increased with task complexity (average correlation: 0.963), with highest differences observed in the PCS task (Fig.~\ref{fig:dialog_gap}a). Smaller language models (e.g., Llama-3.3-70B-Instruct) exhibited higher discrepancies (Fig.~\ref{fig:dialog_gap}a). However, the behavior gap between models narrowed as task complexity increased (Fig.~\ref{fig:dialog_gap}a), suggesting that even advanced models struggle with complex scenarios. Furthermore, we found that LLM agents consistently used more dialog acts per turn compared to humans (Fig.~\ref{fig:dialog_gap}b).

\subsubsection{Tool Use Gap}

To compare the tool usage patterns of human experts and LLM agents, we annotated their turn-by-turn responses using the tool use classifiers detailed in Section~\ref{sec:tool_use}. Similar to dialog acts, the tool use discrepancies, also quantified as $1 - \text{micro-F1 score}$, increased with task complexity (Fig.~\ref{fig:tool_gap}a). Smaller models exhibited higher discrepancies compared to larger models (Fig.~\ref{fig:tool_gap}a), although the gap narrowed for the most complex tasks (Fig.~\ref{fig:tool_gap}a). 

Additionally, LLM agents tend to invoke tools more frequently than human experts (Fig.~\ref{fig:tool_gap}b). This trend was particularly pronounced in smaller models (Fig.~\ref{fig:tool_gap}b), which also exhibited higher discrepancies (Fig.~\ref{fig:tool_gap}a). These observations suggest that LLM agents (particularly smaller models) adopt less efficient and more incorrect tool usage strategies relative to the human benchmark.

\begin{figure}[t]
    \centering
    \subfigure[]{
        \includegraphics[width=0.8\columnwidth]{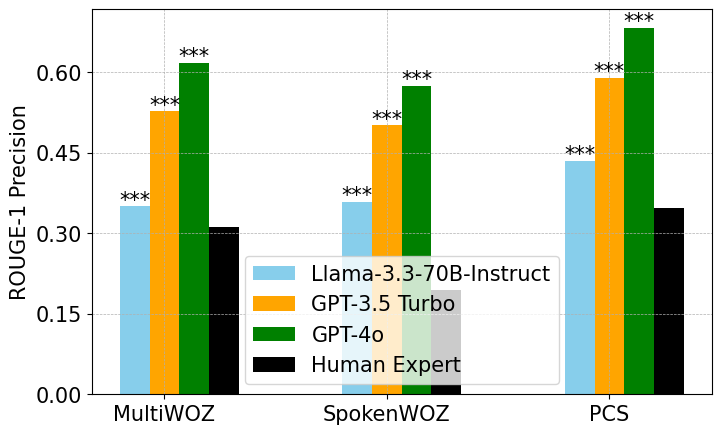}
    }
    \subfigure[]{
        \includegraphics[width=0.8\columnwidth]{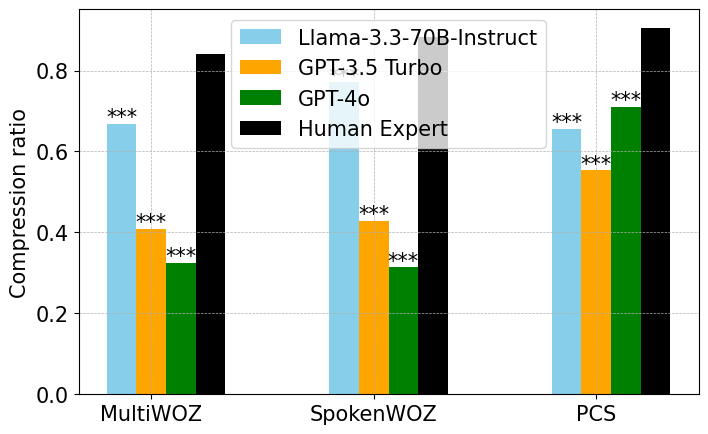}
    }
    \caption{
    \textbf{Knowledge usage gap}. Comparison of knowledge usage patterns between human experts and three LLM agents across tasks. 
    \textbf{(a)} ROUGE-1 precision shows the extent to which agents copy-paste retrieved knowledge into responses. \textbf{(b)} Compression ratio quantifies how efficiently retrieved knowledge is condensed, with higher values indicating greater compression. Asterisks denote statistical significance in the differences between agents and human experts.}
    \label{fig:knowledge_gap}
\end{figure}

\begin{figure}
    \centering
    \includegraphics[width=0.9\linewidth]{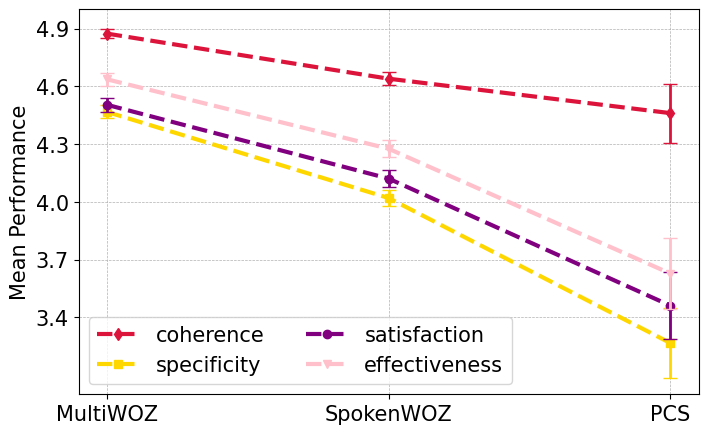}
    \caption{\textbf{Overall performance}. GPT-4o-based agent performance across the four evaluation metrics.
    Higher score indicated better performance. Error bars denote 95\% confidence intervals.}
    \label{fig:perf_vs_complexity}
\end{figure}

\subsubsection{External Knowledge Use Gap}

To compare knowledge usage, we employed two metrics as outlined in Section~\ref{sec:knowledge_use}: ROUGE-1 precision and compression ratio. We found that LLM agents exhibited a higher propensity to verbosely "copy and paste" retrieved knowledge directly into their responses, in contrast to human experts who were more adept at digesting and condensing the knowledge into meaningful insights. This propensity among LLM agents is evidenced by higher ROUGE-1 precision scores (Fig.~\ref{fig:knowledge_gap}a), which indicate the extent of direct copying from retrieved knowledge, and lower compression ratios (Fig.~\ref{fig:knowledge_gap}b), compared to human experts. This behavior persisted even when the agents were explicitly instructed to be concise and provided with concise human responses in the conversation history\footnote{The concise human responses in the conversation history serve as few-shot examples for the LLM agents, but they failed to learn from them.}, as in the teacher-forcing setting. While some degree of copying may occasionally be appropriate—especially for preserving important details—the frequency and extent to which LLM agents rely on this strategy suggest a limited ability to understand and communicate retrieved knowledge effectively.

\subsection{The Impact of Behavior Gap on Task Performance}
We next analyzed how the behavior gap between LLM agents and human experts impacts task performance across tasks of varying complexity. Using the performance evaluator described in Section~\ref{sec:method-performance}, we analyzed the performance for our most advanced model (GPT-4o) and found that the overall performance decreased with increasing task complexity (Fig.~\ref{fig:perf_vs_complexity}), indicating that even state-of-the-art models face challenges with more complex tasks. Below, we explore how the performance relates to the behavior gap (dialog acts and tool usage).

\begin{figure}[t!]
    \centering
    \subfigure[]{
        \includegraphics[width=0.9\columnwidth]{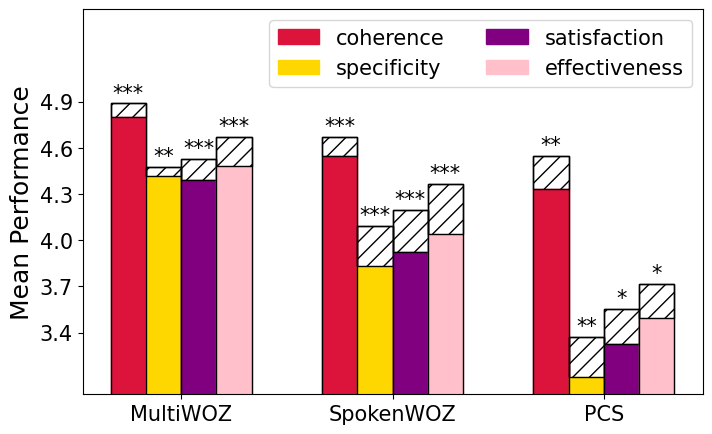}
    }
    \subfigure[]{
        \includegraphics[width=0.9\columnwidth]{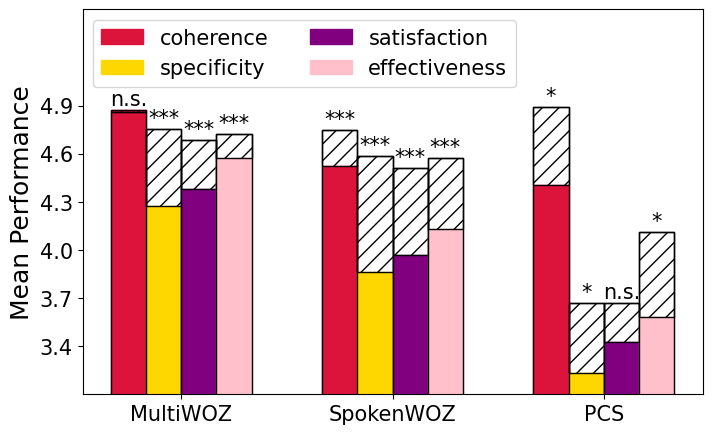}
    }
    \caption{
    \textbf{Correlation of behavior gap with performance}. Comparison of performance on turns with aligned (filled bars) versus misaligned (unfilled bars) dialog acts (a) or tool usage (b). The patterned area shows the performance gap. Asterisks denote statistical significance in the performance differences.}
    \label{fig:performance_impact}
\end{figure}

\paragraph{Statistical Correlation.}
We compared agent performance for turns where human-agent dialog acts and tool usage were aligned (F1 score $\geq 0.5$) versus turns where they were misaligned (F1 score $< 0.5$). The results showed that the response scores were significantly higher ($p < 0.05$) across most metrics and tasks when the behavior was aligned (Fig.~\ref{fig:performance_impact}).

\begin{figure}[t!]
    \centering
    \subfigure[]{
        \includegraphics[width=0.9\columnwidth]{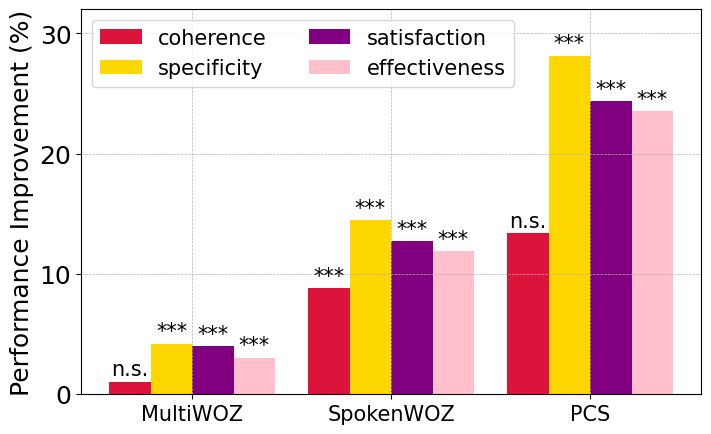}
    }
    \subfigure[]{
        \includegraphics[width=0.9\columnwidth]{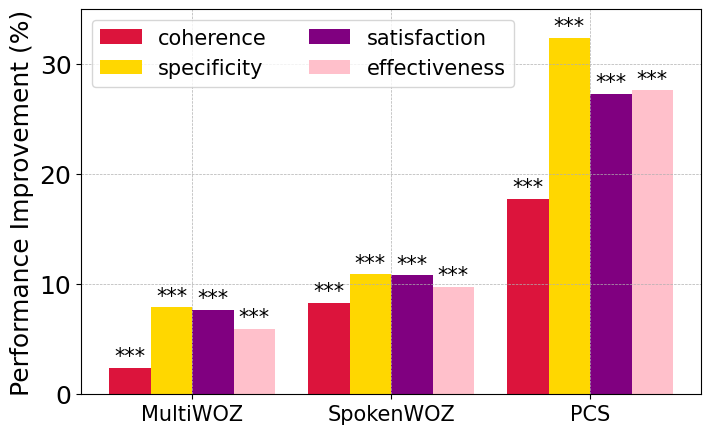}
    }
    \caption{
    \textbf{Improving performance through behavior injection}. 
    Performance improvement (\%) when injecting human dialog acts (a) or human-selected tools (b) into system prompts, compared to the no injection baseline.  Asterisks denote statistical significance in the performance differences.}
    \label{fig:human_injection}
\end{figure}

\paragraph{Behavior Intervention.}
Building on this observation, we tested whether injecting known human dialog acts and selected tools into the system prompt could improve agent performance. Indeed, this intervention significantly improved agent performance across all tasks for most metrics ($p < 0.05$\footnote{The $p$-value corresponds to a one-sample t-test on the log-transformed ratio of performance (injection vs no injection), with the null hypothesis that the log ratio $\leq 0$ (i.e., no improvement).}; Fig.~\ref{fig:human_injection}). This improvement was particularly pronounced in the PCS task (22.4\% and 26.3\% on average for dialog act and tool injection, respectively; Fig.~\ref{fig:human_injection}), highlighting the importance of behavior alignment in handling more complex scenarios.

\section{Discussion}

In this study, we identify a clear behavior gap between LLM agents and human experts% handle conversations 
--a gap that widens as tasks complexity increases. Our analysis reveals three key behavior gaps: LLM agents often adopt different dialog strategies, use tools more frequently but less effectively, and rely on copying retrieved knowledge rather than synthesizing it into meaningful insights. These gaps negatively impact agent performance, but aligning LLM behavior more closely with human strategies mitigates this effect.

Our findings emphasize the importance of enhancing LLM agents' multi-turn dialog capabilities. Our observation that injecting human dialog acts or human-selected tools into prompts improves performance is consistent with prior research on the significance of selecting effective dialog strategies \cite{yu2023prompt, li2024guiding, deng2023plug, zhang_sgp-tod_2023}.  Beyond ensuring smooth dialog flow, addressing gaps in external knowledge usage is crucial, as it enables the agent to efficiently and effectively leverage external knowledge to provide actionable insights.

This requirement is particularly crucial for more complex tasks \citep{zhou2024enhancing, yang2024zhongjing, qian2024chatdev}, as our findings indicate a correlation between task complexity and behavior gap. Collectively, these insights emphasize that training models to grasp not only lexical tokens but also dialog strategies is beneficial.
\section{Limitations}
The accuracy of our analysis depends on the reliability of the LLM-based classifiers used in our framework. While these classifiers were validated on benchmark datasets with ground-truth annotations, applying them to domains that significantly differ from those benchmarks may require further validation to ensure consistent performance.

Moreover, our analysis is confined to turn-level comparisons due to the teacher-forcing evaluation setup. While this setting provides controlled, interpretable diagnostics, it limits our ability to assess full dialog-level strategies and dynamics. Extending the framework to evaluate dialog-level behavior remains an avenue for future work.

Lastly, our study does not include recently released reasoning LLMs such as GPT-o1 or DeepSeek-R1~\cite{guo2025deepseek}, which may exhibit distinct behavioral patterns. Understanding how these newer models behave within our framework is an open question for future exploration.

\bibliography{references}

\begin{thebibliography}{57}
\providecommand{\natexlab}[1]{#1}

\bibitem[{Achiam et~al.(2023)Achiam, Adler, Agarwal, Ahmad, Akkaya, Aleman, Almeida, Altenschmidt, Altman, Anadkat et~al.}]{achiam2023gpt4}
Josh Achiam, Steven Adler, Sandhini Agarwal, Lama Ahmad, Ilge Akkaya, Florencia~Leoni Aleman, Diogo Almeida, Janko Altenschmidt, Sam Altman, Shyamal Anadkat, et~al. 2023.
\newblock Gpt-4 technical report.
\newblock \emph{arXiv preprint arXiv:2303.08774}.

\bibitem[{Adiwardana et~al.(2020)Adiwardana, Luong, So, Hall, Fiedel, Thoppilan, Yang, Kulshreshtha, Nemade, Lu et~al.}]{adiwardana2020towards}
Daniel Adiwardana, Minh-Thang Luong, David~R So, Jamie Hall, Noah Fiedel, Romal Thoppilan, Zi~Yang, Apoorv Kulshreshtha, Gaurav Nemade, Yifeng Lu, et~al. 2020.
\newblock Towards a human-like open-domain chatbot.
\newblock \emph{arXiv preprint arXiv:2001.09977}.

\bibitem[{Austin(1975)}]{austin1975things}
John~Langshaw Austin. 1975.
\newblock \emph{How to do things with words}.
\newblock Harvard university press.

\bibitem[{Bai et~al.(2023)Bai, Bai, Chu, Cui, Dang, Deng, Fan, Ge, Han, Huang et~al.}]{bai2023qwen}
Jinze Bai, Shuai Bai, Yunfei Chu, Zeyu Cui, Kai Dang, Xiaodong Deng, Yang Fan, Wenbin Ge, Yu~Han, Fei Huang, et~al. 2023.
\newblock Qwen technical report.
\newblock \emph{arXiv preprint arXiv:2309.16609}.

\bibitem[{Braggaar et~al.(2024)Braggaar, Liebrecht, Miltenburg, and Krahmer}]{braggaar_evaluating_2024}
Anouck Braggaar, Christine Liebrecht, Emiel~van Miltenburg, and Emiel Krahmer. 2024.
\newblock \href {https://doi.org/10.48550/arXiv.2312.13871} {Evaluating {Task}-oriented {Dialogue} {Systems}: {A} {Systematic} {Review} of {Measures}, {Constructs} and their {Operationalisations}}.
\newblock \emph{arXiv preprint}.
\newblock ArXiv:2312.13871 [cs].

\bibitem[{Brown et~al.(2020)Brown, Mann, Ryder, Subbiah, Kaplan, Dhariwal, Neelakantan, Shyam, Sastry, Askell et~al.}]{brown2020gpt3}
Tom Brown, Benjamin Mann, Nick Ryder, Melanie Subbiah, Jared~D Kaplan, Prafulla Dhariwal, Arvind Neelakantan, Pranav Shyam, Girish Sastry, Amanda Askell, et~al. 2020.
\newblock Language models are few-shot learners.
\newblock \emph{Advances in neural information processing systems}, 33:1877--1901.

\bibitem[{Budzianowski et~al.(2018)Budzianowski, Wen, Tseng, Casanueva, Ultes, Ramadan, and Gasic}]{budzianowski2018multiwoz}
Pawe{\l} Budzianowski, Tsung-Hsien Wen, Bo-Hsiang Tseng, I{\~n}igo Casanueva, Stefan Ultes, Osman Ramadan, and Milica Gasic. 2018.
\newblock Multiwoz-a large-scale multi-domain wizard-of-oz dataset for task-oriented dialogue modelling.
\newblock In \emph{Proceedings of the 2018 Conference on Empirical Methods in Natural Language Processing}, pages 5016--5026.

\bibitem[{Budzianowski et~al.(2020)Budzianowski, Wen, Tseng, Casanueva, Ultes, Ramadan, and Gašić}]{budzianowski_multiwoz_2020}
Paweł Budzianowski, Tsung-Hsien Wen, Bo-Hsiang Tseng, Iñigo Casanueva, Stefan Ultes, Osman Ramadan, and Milica Gašić. 2020.
\newblock \href {https://doi.org/10.48550/arXiv.1810.00278} {{MultiWOZ} -- {A} {Large}-{Scale} {Multi}-{Domain} {Wizard}-of-{Oz} {Dataset} for {Task}-{Oriented} {Dialogue} {Modelling}}.
\newblock \emph{arXiv preprint}.
\newblock ArXiv:1810.00278 [cs].

\bibitem[{Bunt(2012)}]{bunt2012dynamic}
Harry Bunt. 2012.
\newblock Dynamic interpretation and dialogue theory.
\newblock In \emph{The structure of multimodal dialogue II}, pages 139--166. John Benjamins Publishing Company.

\bibitem[{Bunt et~al.(2012)Bunt, Alexandersson, Choe, Fang, Hasida, Petukhova, Popescu-Belis, and Traum}]{bunt2012iso}
Harry Bunt, Jan Alexandersson, Jae-Woong Choe, Alex~Chengyu Fang, Koiti Hasida, Volha Petukhova, Andrei Popescu-Belis, and David~R Traum. 2012.
\newblock Iso 24617-2: A semantically-based standard for dialogue annotation.
\newblock In \emph{LREC}, pages 430--437.

\bibitem[{Bunt et~al.(2019)Bunt, Petukhova, Malchanau, Fang, and Wijnhoven}]{bunt2019dialogbank}
Harry Bunt, Volha Petukhova, Andrei Malchanau, Alex Fang, and Kars Wijnhoven. 2019.
\newblock The dialogbank: dialogues with interoperable annotations.
\newblock \emph{Language Resources and Evaluation}, 53:213--249.

\bibitem[{Deng et~al.(2023)Deng, Zhang, Lam, Ng, and Chua}]{deng2023plug}
Yang Deng, Wenxuan Zhang, Wai Lam, See-Kiong Ng, and Tat-Seng Chua. 2023.
\newblock Plug-and-play policy planner for large language model powered dialogue agents.
\newblock In \emph{The Twelfth International Conference on Learning Representations}.

\bibitem[{Dong et~al.(2025)Dong, Chen, and Yang}]{dong2025protod}
Wenjie Dong, Sirong Chen, and Yan Yang. 2025.
\newblock Protod: Proactive task-oriented dialogue system based on large language model.
\newblock In \emph{Proceedings of the 31st International Conference on Computational Linguistics}, pages 9147--9164.

\bibitem[{Dubey et~al.(2024)Dubey, Jauhri, Pandey, Kadian, Al-Dahle, Letman, Mathur, Schelten, Yang, Fan et~al.}]{dubey2024llama}
Abhimanyu Dubey, Abhinav Jauhri, Abhinav Pandey, Abhishek Kadian, Ahmad Al-Dahle, Aiesha Letman, Akhil Mathur, Alan Schelten, Amy Yang, Angela Fan, et~al. 2024.
\newblock The llama 3 herd of models.
\newblock \emph{arXiv preprint arXiv:2407.21783}.

\bibitem[{Elizabeth et~al.(2024)Elizabeth, Veyret, Couceiro, Dusek, and Rojas-Barahona}]{elizabeth2024large}
Michelle Elizabeth, Morgan Veyret, Miguel Couceiro, Ondrej Dusek, and Lina~M Rojas-Barahona. 2024.
\newblock Do large language models with reasoning and acting meet the needs of task-oriented dialogue?
\newblock \emph{arXiv preprint arXiv:2412.01262}.

\bibitem[{Feng et~al.(2023)Feng, Jiao, Prasad, Aletras, Yilmaz, and Kazai}]{feng2023schema}
Yue Feng, Yunlong Jiao, Animesh Prasad, Nikolaos Aletras, Emine Yilmaz, and Gabriella Kazai. 2023.
\newblock Schema-guided user satisfaction modeling for task-oriented dialogues.
\newblock \emph{arXiv preprint arXiv:2305.16798}.

\bibitem[{Guo et~al.(2025)Guo, Yang, Zhang, Song, Zhang, Xu, Zhu, Ma, Wang, Bi et~al.}]{guo2025deepseek}
Daya Guo, Dejian Yang, Haowei Zhang, Junxiao Song, Ruoyu Zhang, Runxin Xu, Qihao Zhu, Shirong Ma, Peiyi Wang, Xiao Bi, et~al. 2025.
\newblock Deepseek-r1: Incentivizing reasoning capability in llms via reinforcement learning.
\newblock \emph{arXiv preprint arXiv:2501.12948}.

\bibitem[{Guo et~al.(2018)Guo, Wu, Cheng, Rennie, Tesauro, and Feris}]{guo2018dialog}
Xiaoxiao Guo, Hui Wu, Yu~Cheng, Steven Rennie, Gerald Tesauro, and Rogerio Feris. 2018.
\newblock Dialog-based interactive image retrieval.
\newblock \emph{Advances in neural information processing systems}, 31.

\bibitem[{Gupta et~al.(2024)Gupta, Ravichandran, Zhang, Shah, Beniwal, and Sadagopan}]{gupta2024dard}
Aman Gupta, Anirudh Ravichandran, Ziji Zhang, Swair Shah, Anurag Beniwal, and Narayanan Sadagopan. 2024.
\newblock Dard: A multi-agent approach for task-oriented dialog systems.
\newblock \emph{arXiv preprint arXiv:2411.00427}.

\bibitem[{Heck et~al.(2023)Heck, Lubis, Ruppik, Vukovic, Feng, Geishauser, Lin, van Niekerk, and Gasic}]{heck2023chatgpt}
Michael Heck, Nurul Lubis, Benjamin~Matthias Ruppik, Renato Vukovic, Shutong Feng, Christian Geishauser, Hsien-chin Lin, Carel van Niekerk, and Milica Gasic. 2023.
\newblock Chatgpt for zero-shot dialogue state tracking: A solution or an opportunity?
\newblock In \emph{The 61st Annual Meeting Of The Association For Computational Linguistics}.

\bibitem[{Hudeček and Dušek(2023)}]{hudecek_are_2023}
Vojtěch Hudeček and Ondřej Dušek. 2023.
\newblock \href {https://doi.org/10.48550/arXiv.2304.06556} {Are {LLMs} {All} {You} {Need} for {Task}-{Oriented} {Dialogue}?}
\newblock \emph{arXiv preprint}.
\newblock ArXiv:2304.06556 [cs].

\bibitem[{Hurst et~al.(2024)Hurst, Lerer, Goucher, Perelman, Ramesh, Clark, Ostrow, Welihinda, Hayes, Radford et~al.}]{hurst2024gpt}
Aaron Hurst, Adam Lerer, Adam~P Goucher, Adam Perelman, Aditya Ramesh, Aidan Clark, AJ~Ostrow, Akila Welihinda, Alan Hayes, Alec Radford, et~al. 2024.
\newblock Gpt-4o system card.
\newblock \emph{arXiv preprint arXiv:2410.21276}.

\bibitem[{Jiang et~al.(2023)Jiang, Sablayrolles, Mensch, Bamford, Chaplot, Casas, Bressand, Lengyel, Lample, Saulnier et~al.}]{jiang2023mistral}
Albert~Q Jiang, Alexandre Sablayrolles, Arthur Mensch, Chris Bamford, Devendra~Singh Chaplot, Diego de~las Casas, Florian Bressand, Gianna Lengyel, Guillaume Lample, Lucile Saulnier, et~al. 2023.
\newblock Mistral 7b.
\newblock \emph{arXiv preprint arXiv:2310.06825}.

\bibitem[{{LangChain Inc.}(2024)}]{langgraph2024}
{LangChain Inc.} 2024.
\newblock \href {https://langchain-ai.github.io/langgraph/} {Langgraph: A library for building stateful, multi-actor applications with llms}.
\newblock LangChain Documentation.

\bibitem[{Lewis et~al.(2020)Lewis, Perez, Piktus, Petroni, Karpukhin, Goyal, K{\"u}ttler, Lewis, Yih, Rockt{\"a}schel et~al.}]{lewis2020retrieval}
Patrick Lewis, Ethan Perez, Aleksandra Piktus, Fabio Petroni, Vladimir Karpukhin, Naman Goyal, Heinrich K{\"u}ttler, Mike Lewis, Wen-tau Yih, Tim Rockt{\"a}schel, et~al. 2020.
\newblock Retrieval-augmented generation for knowledge-intensive nlp tasks.
\newblock \emph{Advances in Neural Information Processing Systems}, 33:9459--9474.

\bibitem[{Li et~al.(2017)Li, Chen, Li, Gao, and Celikyilmaz}]{li2017end}
Xiujun Li, Yun-Nung Chen, Lihong Li, Jianfeng Gao, and Asli Celikyilmaz. 2017.
\newblock End-to-end task-completion neural dialogue systems.
\newblock \emph{arXiv preprint arXiv:1703.01008}.

\bibitem[{Li et~al.(2016)Li, Lipton, Dhingra, Li, Gao, and Chen}]{li2016user}
Xiujun Li, Zachary~C Lipton, Bhuwan Dhingra, Lihong Li, Jianfeng Gao, and Yun-Nung Chen. 2016.
\newblock A user simulator for task-completion dialogues.
\newblock \emph{arXiv preprint arXiv:1612.05688}.

\bibitem[{Li et~al.(2024)Li, Peng, He, Galley, Gao, and Yan}]{li2024guiding}
Zekun Li, Baolin Peng, Pengcheng He, Michel Galley, Jianfeng Gao, and Xifeng Yan. 2024.
\newblock Guiding large language models via directional stimulus prompting.
\newblock \emph{Advances in Neural Information Processing Systems}, 36.

\bibitem[{Liu et~al.(2023)Liu, Iter, Xu, Wang, Xu, and Zhu}]{liu2023g}
Yang Liu, Dan Iter, Yichong Xu, Shuohang Wang, Ruochen Xu, and Chenguang Zhu. 2023.
\newblock G-eval: Nlg evaluation using gpt-4 with better human alignment.
\newblock In \emph{Proceedings of the 2023 Conference on Empirical Methods in Natural Language Processing}, pages 2511--2522.

\bibitem[{Mehri and Eskenazi(2020)}]{mehri2020fed}
Shikib Mehri and Maxine Eskenazi. 2020.
\newblock Unsupervised evaluation of interactive dialog with dialogpt.
\newblock In \emph{Proceedings of the 21th Annual Meeting of the Special Interest Group on Discourse and Dialogue}, pages 225--235.

\bibitem[{Mezza et~al.(2018)Mezza, Cervone, Tortoreto, Stepanov, Riccardi et~al.}]{mezza2018iso}
Stefano Mezza, Alessandra Cervone, Giuliano Tortoreto, Evgeny~A Stepanov, Giuseppe Riccardi, et~al. 2018.
\newblock Iso-standard domain-independent dialogue act tagging for conversational agents.
\newblock In \emph{Proceedings of the 27th International Conference on Computational Linguistics}, pages 3539--3551. Association for Computational Linguistics.

\bibitem[{Nekvinda and Dušek(2021)}]{nekvinda_shades_2021}
Tomáš Nekvinda and Ondřej Dušek. 2021.
\newblock \href {https://doi.org/10.48550/arXiv.2106.05555} {Shades of {BLEU}, {Flavours} of {Success}: {The} {Case} of {MultiWOZ}}.
\newblock \emph{arXiv preprint}.
\newblock ArXiv:2106.05555 [cs].

\bibitem[{Ouyang et~al.(2022)Ouyang, Wu, Jiang, Almeida, Wainwright, Mishkin, Zhang, Agarwal, Slama, Ray et~al.}]{ouyang2022training}
Long Ouyang, Jeffrey Wu, Xu~Jiang, Diogo Almeida, Carroll Wainwright, Pamela Mishkin, Chong Zhang, Sandhini Agarwal, Katarina Slama, Alex Ray, et~al. 2022.
\newblock Training language models to follow instructions with human feedback.
\newblock \emph{Advances in neural information processing systems}, 35:27730--27744.

\bibitem[{Qian et~al.(2024)Qian, Liu, Liu, Chen, Dang, Li, Yang, Chen, Su, Cong et~al.}]{qian2024chatdev}
Chen Qian, Wei Liu, Hongzhang Liu, Nuo Chen, Yufan Dang, Jiahao Li, Cheng Yang, Weize Chen, Yusheng Su, Xin Cong, et~al. 2024.
\newblock Chatdev: Communicative agents for software development.
\newblock In \emph{Proceedings of the 62nd Annual Meeting of the Association for Computational Linguistics (Volume 1: Long Papers)}, pages 15174--15186.

\bibitem[{Qin et~al.(2024)Qin, Hu, Lin, Chen, Ding, Cui, Zeng, Zhou, Huang, Xiao et~al.}]{qin2024tool}
Yujia Qin, Shengding Hu, Yankai Lin, Weize Chen, Ning Ding, Ganqu Cui, Zheni Zeng, Xuanhe Zhou, Yufei Huang, Chaojun Xiao, et~al. 2024.
\newblock Tool learning with foundation models.
\newblock \emph{ACM Computing Surveys}, 57(4):1--40.

\bibitem[{Rastogi et~al.(2020)Rastogi, Zang, Sunkara, Gupta, and Khaitan}]{rastogi_towards_2020}
Abhinav Rastogi, Xiaoxue Zang, Srinivas Sunkara, Raghav Gupta, and Pranav Khaitan. 2020.
\newblock \href {https://doi.org/10.48550/arXiv.1909.05855} {Towards {Scalable} {Multi}-domain {Conversational} {Agents}: {The} {Schema}-{Guided} {Dialogue} {Dataset}}.
\newblock \emph{arXiv preprint}.
\newblock ArXiv:1909.05855 [cs].

\bibitem[{Shaikh et~al.(2024)Shaikh, Gligori{\'c}, Khetan, Gerstgrasser, Yang, and Jurafsky}]{shaikh2024grounding}
Omar Shaikh, Kristina Gligori{\'c}, Ashna Khetan, Matthias Gerstgrasser, Diyi Yang, and Dan Jurafsky. 2024.
\newblock Grounding gaps in language model generations.
\newblock In \emph{Proceedings of the 2024 Conference of the North American Chapter of the Association for Computational Linguistics: Human Language Technologies (Volume 1: Long Papers)}, pages 6279--6296.

\bibitem[{Si et~al.(2024)Si, Ma, Gao, Wu, Lin, Dai, Li, Yan, Huang, and Li}]{si2024spokenwoz}
Shuzheng Si, Wentao Ma, Haoyu Gao, Yuchuan Wu, Ting-En Lin, Yinpei Dai, Hangyu Li, Rui Yan, Fei Huang, and Yongbin Li. 2024.
\newblock Spokenwoz: A large-scale speech-text benchmark for spoken task-oriented dialogue agents.
\newblock \emph{Advances in Neural Information Processing Systems}, 36.

\bibitem[{Stolcke et~al.(2000)Stolcke, Ries, Coccaro, Shriberg, Bates, Jurafsky, Taylor, Martin, Ess-Dykema, and Meteer}]{stolcke2000dialogue}
Andreas Stolcke, Klaus Ries, Noah Coccaro, Elizabeth Shriberg, Rebecca Bates, Daniel Jurafsky, Paul Taylor, Rachel Martin, Carol~Van Ess-Dykema, and Marie Meteer. 2000.
\newblock Dialogue act modeling for automatic tagging and recognition of conversational speech.
\newblock \emph{Computational linguistics}, 26(3):339--373.

\bibitem[{Team et~al.(2023)Team, Anil, Borgeaud, Alayrac, Yu, Soricut, Schalkwyk, Dai, Hauth, Millican et~al.}]{team2023gemini}
Gemini Team, Rohan Anil, Sebastian Borgeaud, Jean-Baptiste Alayrac, Jiahui Yu, Radu Soricut, Johan Schalkwyk, Andrew~M Dai, Anja Hauth, Katie Millican, et~al. 2023.
\newblock Gemini: a family of highly capable multimodal models.
\newblock \emph{arXiv preprint arXiv:2312.11805}.

\bibitem[{Touvron et~al.(2023)Touvron, Lavril, Izacard, Martinet, Lachaux, Lacroix, Rozi{\`e}re, Goyal, Hambro, Azhar et~al.}]{touvron2023llama}
Hugo Touvron, Thibaut Lavril, Gautier Izacard, Xavier Martinet, Marie-Anne Lachaux, Timoth{\'e}e Lacroix, Baptiste Rozi{\`e}re, Naman Goyal, Eric Hambro, Faisal Azhar, et~al. 2023.
\newblock Llama: Open and efficient foundation language models.
\newblock \emph{arXiv preprint arXiv:2302.13971}.

\bibitem[{Venkatesh et~al.(2018)Venkatesh, Khatri, Ram, Guo, Gabriel, Nagar, Prasad, Cheng, Hedayatnia, Metallinou et~al.}]{venkatesh2018evaluating}
Anu Venkatesh, Chandra Khatri, Ashwin Ram, Fenfei Guo, Raefer Gabriel, Ashish Nagar, Rohit Prasad, Ming Cheng, Behnam Hedayatnia, Angeliki Metallinou, et~al. 2018.
\newblock On evaluating and comparing open domain dialog systems.
\newblock \emph{arXiv preprint arXiv:1801.03625}.

\bibitem[{Wang et~al.(2024)Wang, Ma, Feng, Zhang, Yang, Zhang, Chen, Tang, Chen, Lin et~al.}]{wang2024survey}
Lei Wang, Chen Ma, Xueyang Feng, Zeyu Zhang, Hao Yang, Jingsen Zhang, Zhiyuan Chen, Jiakai Tang, Xu~Chen, Yankai Lin, et~al. 2024.
\newblock A survey on large language model based autonomous agents.
\newblock \emph{Frontiers of Computer Science}, 18(6).

\bibitem[{Williams and Zipser(1989)}]{williams1989learning}
Ronald~J Williams and David Zipser. 1989.
\newblock A learning algorithm for continually running fully recurrent neural networks.
\newblock \emph{Neural computation}, 1(2):270--280.

\bibitem[{Xi et~al.(2025)Xi, Chen, Guo, He, Ding, Hong, Zhang, Wang, Jin, Zhou et~al.}]{xi2025rise}
Zhiheng Xi, Wenxiang Chen, Xin Guo, Wei He, Yiwen Ding, Boyang Hong, Ming Zhang, Junzhe Wang, Senjie Jin, Enyu Zhou, et~al. 2025.
\newblock The rise and potential of large language model based agents: A survey.
\newblock \emph{Science China Information Sciences}, 68(2):121101.

\bibitem[{Xu et~al.(2024)Xu, Mao, Yang, Sun, and Huang}]{xu2024rethinking}
Heng-Da Xu, Xian-Ling Mao, Puhai Yang, Fanshu Sun, and He-Yan Huang. 2024.
\newblock Rethinking task-oriented dialogue systems: From complex modularity to zero-shot autonomous agent.
\newblock In \emph{Proceedings of the 62nd Annual Meeting of the Association for Computational Linguistics (Volume 1: Long Papers)}, pages 2748--2763.

\bibitem[{Yang et~al.(2024)Yang, Zhao, Zhu, Zhou, Xu, Jia, and Zan}]{yang2024zhongjing}
Songhua Yang, Hanjie Zhao, Senbin Zhu, Guangyu Zhou, Hongfei Xu, Yuxiang Jia, and Hongying Zan. 2024.
\newblock Zhongjing: Enhancing the chinese medical capabilities of large language model through expert feedback and real-world multi-turn dialogue.
\newblock In \emph{Proceedings of the AAAI Conference on Artificial Intelligence}, volume~38, pages 19368--19376.

\bibitem[{Yao et~al.(2022)Yao, Zhao, Yu, Du, Shafran, Narasimhan, and Cao}]{yaoreact}
Shunyu Yao, Jeffrey Zhao, Dian Yu, Nan Du, Izhak Shafran, Karthik~R Narasimhan, and Yuan Cao. 2022.
\newblock React: Synergizing reasoning and acting in language models.
\newblock In \emph{The Eleventh International Conference on Learning Representations}.

\bibitem[{Yu et~al.(2023)Yu, Chen, and Yu}]{yu2023prompt}
Xiao Yu, Maximillian Chen, and Zhou Yu. 2023.
\newblock Prompt-based monte-carlo tree search for goal-oriented dialogue policy planning.
\newblock In \emph{Proceedings of the 2023 Conference on Empirical Methods in Natural Language Processing}, pages 7101--7125.

\bibitem[{Zang et~al.(2020)Zang, Rastogi, Sunkara, Gupta, Zhang, and Chen}]{zang_multiwoz_2020}
Xiaoxue Zang, Abhinav Rastogi, Srinivas Sunkara, Raghav Gupta, Jianguo Zhang, and Jindong Chen. 2020.
\newblock \href {https://doi.org/10.48550/arXiv.2007.12720} {{MultiWOZ} 2.2 : {A} {Dialogue} {Dataset} with {Additional} {Annotation} {Corrections} and {State} {Tracking} {Baselines}}.
\newblock \emph{arXiv preprint}.
\newblock ArXiv:2007.12720 [cs].

\bibitem[{Zhang et~al.(2023{\natexlab{a}})Zhang, Peng, Li, Zhou, and Meng}]{zhang2023sgp}
Xiaoying Zhang, Baolin Peng, Kun Li, Jingyan Zhou, and Helen Meng. 2023{\natexlab{a}}.
\newblock Sgp-tod: Building task bots effortlessly via schema-guided llm prompting.
\newblock In \emph{Findings of the Association for Computational Linguistics: EMNLP 2023}, pages 13348--13369.

\bibitem[{Zhang et~al.(2023{\natexlab{b}})Zhang, Peng, Li, Zhou, and Meng}]{zhang_sgp-tod_2023}
Xiaoying Zhang, Baolin Peng, Kun Li, Jingyan Zhou, and Helen Meng. 2023{\natexlab{b}}.
\newblock \href {https://doi.org/10.18653/v1/2023.findings-emnlp.891} {{SGP}-{TOD}: {Building} {Task} {Bots} {Effortlessly} via {Schema}-{Guided} {LLM} {Prompting}}.
\newblock In \emph{Findings of the {Association} for {Computational} {Linguistics}: {EMNLP} 2023}, pages 13348--13369, Singapore. Association for Computational Linguistics.

\bibitem[{Zhang et~al.(2020)Zhang, Sun, Galley, Chen, Brockett, Gao, Gao, Liu, and Dolan}]{zhang2020dialogpt}
Yizhe Zhang, Siqi Sun, Michel Galley, Yen-Chun Chen, Chris Brockett, Xiang Gao, Jianfeng Gao, Jingjing Liu, and William~B Dolan. 2020.
\newblock Dialogpt: Large-scale generative pre-training for conversational response generation.
\newblock In \emph{Proceedings of the 58th Annual Meeting of the Association for Computational Linguistics: System Demonstrations}, pages 270--278.

\bibitem[{Zhang et~al.(2022)Zhang, Sun, Gao, Fang, Brockett, Galley, Gao, and Dolan}]{zhang2022retgen}
Yizhe Zhang, Siqi Sun, Xiang Gao, Yuwei Fang, Chris Brockett, Michel Galley, Jianfeng Gao, and Bill Dolan. 2022.
\newblock Retgen: A joint framework for retrieval and grounded text generation modeling.
\newblock In \emph{Proceedings of the AAAI Conference on Artificial Intelligence}, volume~36, pages 11739--11747.

\bibitem[{Zhao et~al.(2024)Zhao, Huang, Xu, Lin, Liu, and Huang}]{zhao2024expel}
Andrew Zhao, Daniel Huang, Quentin Xu, Matthieu Lin, Yong-Jin Liu, and Gao Huang. 2024.
\newblock Expel: Llm agents are experiential learners.
\newblock In \emph{Proceedings of the AAAI Conference on Artificial Intelligence}, volume~38, pages 19632--19642.

\bibitem[{Zheng et~al.(2023)Zheng, Chiang, Sheng, Zhuang, Wu, Zhuang, Lin, Li, Li, Xing et~al.}]{zheng2023judging}
Lianmin Zheng, Wei-Lin Chiang, Ying Sheng, Siyuan Zhuang, Zhanghao Wu, Yonghao Zhuang, Zi~Lin, Zhuohan Li, Dacheng Li, Eric Xing, et~al. 2023.
\newblock Judging llm-as-a-judge with mt-bench and chatbot arena.
\newblock \emph{Advances in Neural Information Processing Systems}, 36:46595--46623.

\bibitem[{Zhou et~al.(2024)Zhou, Zhang, Zhang, and Yuan}]{zhou2024enhancing}
Jiahao Zhou, Qiang Zhang, Fengda Zhang, and Caixia Yuan. 2024.
\newblock Enhancing troubleshooting task-oriented dialog systems with large language models.
\newblock In \emph{International Conference on Intelligent Robotics and Applications}, pages 328--338. Springer.

\end{thebibliography}

\appendix

\section{Appendix}
\label{sec:appendix}

\subsection{Datasets}
\label{sec:appendix_datasets}

\paragraph{MultiWOZ.}
The Multi-Domain Wizard-of-Oz (MultiWOZ) dataset~\cite{budzianowski2018multiwoz} is a widely used benchmark for evaluating TOD systems. It consists of over 10,000 human-human written dialogs spanning seven domains: Attraction, Hospital, Police, Hotel, Restaurant, Taxi, and Train. Its text-based modality and low average number of turns per dialog make it the simplest among the three datasets in terms of complexity. 

For our experiments, we specifically utilized the test set of MultiWOZ 2.2~\cite{zang_multiwoz_2020}, which includes 1,000 dialogs (Table\ref{tab:dataset}) across five domains: hotel, attraction, restaurant, taxi, and train. 

\paragraph{SpokenWOZ.}
SpokenWOZ~\cite{si2024spokenwoz} builds upon MultiWOZ by introducing challenges associated with spoken language. It was constructed from human-to-human spoken conversations conducted via phone calls, transcribed to text. This introduces additional complexity due to unique challenges inherent in spoken conversation such as incomplete utterances. The average number of turns per dialog increases to 36~(Table~\ref{tab:dataset}), reflecting the more verbose nature of spoken interactions compared to written ones. 

For our study, we focused on a subset of the SpokenWOZ test set that aligns with the five domains used in MultiWOZ: hotel, attraction, restaurant, taxi, and train. This resulted in a test set containing 987 dialogs (Table~\ref{tab:dataset}) .

\paragraph{PCS.}
The dialogs are filtered based on the impact of the agent's response on user satisfaction, as judged by GPT-4o, so the human agent responses can serve as a proxy for high-quality standards.
This dataset contained 832 dialogs (Table~\ref{tab:dataset}) with an average of 120 turns per dialog (Table~\ref{tab:dataset}).
We present a snippet of the PCS dataset in Table~\ref{tab:pcs_snippet}.

\begin{table}
    \centering
    \small
    \begin{tabular}{l p{6cm}}
        \toprule
         speaker & content \\
          \hline
        agent & Thank you for calling [PRODUCT NAME]. My name is [PII], can I have your name, please? \\
        customer & My name is [PII] \\
        agent & Hi [PII], how are you doing today? \\
        customer & I'm doing fine, thank you. \\
        agent & How can I help you? [PII]. \\
        customer & I have [PRODUCT NAME]. And I'm trying to, you know, I've downloaded transactions and I would like to deposit them and match them with the transaction's, but I'm doing something wrong because it seems to take the deposit's but it's not matching with the downloads. \\
        agent & Hello. I see. And what is the year version of the [PRODUCT NAME] you have? \\
        customer & It is [VERSION] [PRODUCT NAME]. \\
        agent & All right. And what do you mean by it's not matching with the transactions that you have? So you enter, have you entered transactions manually inside [PRODUCT NAME] and then you download the transactions on your bank? \\
        \bottomrule
    \end{tabular}
    \caption{A snippet for the PCS task.} 
    \label{tab:pcs_snippet}
\end{table}

\subsection{Zero-Shot Agent}
\label{sec:appendix_agents}

Here we provide details on the the Zero-shot ReAct agent employed in this work. See Table.~\ref{tab:agent_system_prompt} for a sample system prompt.

\subsubsection{Tool and Prompt Configurations}

\paragraph{MultiWOZ Task.}
For the MultiWOZ dataset, the agent was equipped with tools to allow information retrieval and booking across five domains: \textit{hotel}, \textit{attraction}, \textit{restaurant}, \textit{taxi}, and \textit{train}. The following tools were integrated into the agent:
\begin{itemize}
    \item \textbf{Search Tools:} These tools enabled the agent to query the MultiWOZ database to retrieve information about hotels (\texttt{FindHotels}), attractions (\texttt{FindAttractions}), restaurants (\texttt{FindRestaurants}), and trains (\texttt{FindTrains}) based on user preferences (e.g., location, price range).
    \item \textbf{Booking Tools:} These tools allowed the agent to book hotels (\texttt{BookHotel}), restaurants (\texttt{BookRestaurant}), trains (\texttt{BookTrain}), and taxis (\texttt{BookTaxi}).
\end{itemize}

To minimize hallucination and ensure accurate tool usage, we provided the agent with an exhaustive list of all entities present in the MultiWOZ database as part of its system prompt. This ensured that the agent could accurately identify valid entities during search and booking operations. 

\paragraph{SpokenWOZ Task.}
The SpokenWOZ task builds upon MultiWOZ by introducing additional complexity due to its spoken modality. The agent for this task was configured with all the tools available in MultiWOZ (search and booking tools) along with an additional tool:
\begin{itemize}
    \item \texttt{BookParking}: This tool enabled the agent to book parking spaces for users when requested.
\end{itemize}

Similar to MultiWOZ, an exhaustive list of entities was included in the system prompt to reduce hallucination and ensure accurate tool usage. Additionally, once the user’s booking is successful, the agent was prompted to not only provide the entity booked but also ask for the user’s profile information such as name, ID, email, license plate number, and phone.

\paragraph{PCS Task.}
The PCS dataset presented unique challenges due to its real-world nature and diverse task requirements.  Unlike traditional slot-filling tasks, this task does not have a predefined list of slots due to significant variations in individual user circumstances and the inherent complexity of the service. 
The complexity also comes from the spoken characteristics, such as incomplete utterances and back-channel. For this dataset, we equipped the agent with four specialized tools based on tool usage pattern from the human experts:
\begin{itemize}
    \item \texttt{KnowledgeLookup}: This tool allowed the agent to search an internal knowledge base as well as the internet for product-related information.
    \item \texttt{CustomerInfoLookup}: This tool simulated the agent retrieving customer-specific information.
    \item \texttt{EscalateOrTransfer}: This tool simulated the agent escalating unresolved issues to other teams or departments when necessary.
    \item \texttt{ScreenShare}: This tool simulated initiating a screen-sharing session with the customer.
\end{itemize}

The \texttt{Customer Info Lookup, Escalation}, and \texttt{ScreenShare} tools did not return specific information but instead provided confirmation of their usage. This design ensured that the behavior analysis focused on when and how tools were invoked rather than the detailed content of their outputs. The \texttt{KnowledgeLookup} tool, in contrast, was fully implemented returning detailed responses containing information such as relevant product information, troubleshooting guides, etc.

\subsection{Task-Specific WOZ Framework}
\label{sec:appendix_woz_framework}
The task-specific WOZ framework was based on the dialogue act types defined in the MultiWOZ 2.2 dataset~\cite{zang_multiwoz_2020}. This framework was only used
for analyzing the agent's performance on the MultiWOZ and SpokenWOZ datasets. 
It included 10 dialogue act types: \texttt{inform, request, select, recommend, nooffer, offerbook, book, nobook, greet}, and \texttt{reqmore}. These acts combined capture key conversational behaviors in the MultiWOZ and SpokenWOZ datasets.

To ensure clarity and avoid overlapping definitions that might confuse our LLM-based classifier, we merged certain act types that were originally present in the MultiWOZ 2.2 dataset. More specifically:
\begin{itemize}
    \item The \texttt{greet} act combined similar acts such as "welcome," "greet," "bye," and "thanks" from MultiWOZ into a single unified category.
    \item The \texttt{book} act combined both "book" and "offerbooked" into one category since they were closely related in function.
\end{itemize}

These modifications simplified the annotation process while preserving the semantic distinctions necessary for analyzing dialogue actions.

\subsection{LLM-Based Dialog classifiers}
\label{sec:appendix_dialog_classifier}
Both the classifiers were designed to receive the user input and the corresponding agent response as input and output a list of predicted dialogue act types present in the agent's response. The classifiers also provide reasoning for the presence or absence of each act, which improved its accuracy. We implemented structured outputs to ensure consistency and robustness. in the classifier's outputs. 

The system prompt for both the classifiers included definitions of each dialogue act type. It also included few-shot examples for each of the dialogue act types. See Table.~\ref{tab:dialog_act_prompt} for a sample system prompt.

Each of our classifier was validated against ground-truth annotations from established datasets:
\begin{enumerate}
    \item WOZ classifier: For this classifier, we validated predictions using the annotated dialogue acts from the test set of MultiWOZ 2.2~\cite{budzianowski_multiwoz_2020} dataset. 
    \item ISO classifier: For this classifier, we validated predictions using the DialogBank dataset~\cite{bunt2019dialogbank}, which contains dialogues from various sources annotated according to the ISO 24617-2 standards.
\end{enumerate}

\subsection{Tool Classifier}
\label{sec:appendix_tool_classifier}

We developed task-specific tool classifiers for each of the three tasks: MultiWOZ, SpokenWOZ, and PCS. Similar to the dialogue act classifiers, the tool usage classifiers were designed to take the user input and the corresponding agent response as input and output a list of predicted tools used during the generation of the agent's response. Additionally, the classifiers also generated reasoning for the presence or absence of each tool in the agent's response, and employed structured outputs to ensure consistency and robustness in the classifier's outputs. 

The system prompts included tool definitions tailored to each task (see Table~\ref{tab:tool_use_prompt} for a sample prompt and Section~\ref{sec:appendix_agents} for task-specific tool lists).

\subsection{Performance evaluator}
\label{sec:appendix_perf_eval}
\begin{figure}
    \centering
    \includegraphics[width=0.95\linewidth]{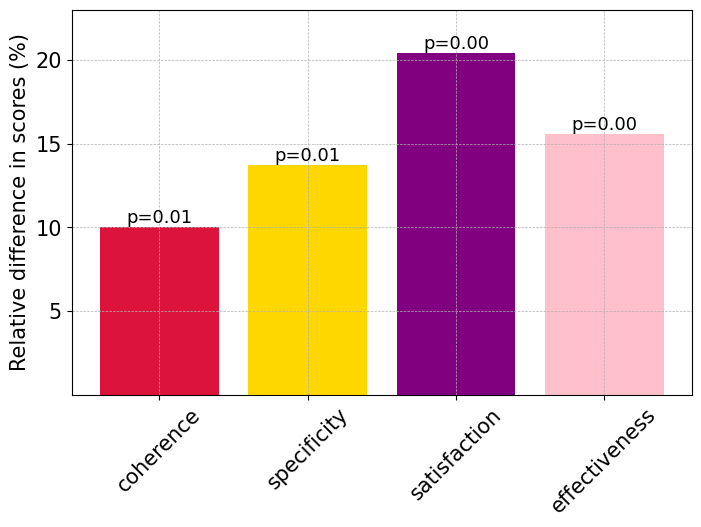}
    \caption{Relative difference (\%) in scores for the four metrics for dialogs with success rate of 1 versus dialogs with success rate of 0.}
    \label{fig:perf_validation}
\end{figure}

To validate the response quality evaluator, we used the test set of the MultiWOZ 2.2~\cite{budzianowski_multiwoz_2020} dataset. We compared the aggregated predicted scores for agent responses in each dialog from the validation dataset with the success rate for those dialogs, calculated using the method outlined by \citet{nekvinda_shades_2021}. The validation results are presented in Figure~\ref{fig:perf_validation}.

\begin{table*}[t]
\small
\begin{tabular}{|p{0.95\textwidth}|}
\hline
\begin{verbatim}
***Task Description***
You are a multi-domain customer support assistant for booking various services in Cambridge, UK, 
including hotel bookings, restaurant reservations, train bookings, taxi bookings, and finding 
attractions. All the services and information you provide are specific to Cambridge, UK and its
surrounding areas.

***Database Details***
Available options in the database:
Areas in Cambridge: west, centre, south, east, north
Train Departure Stations: leicester, kings lynn, cambridge, ...
Train Arrival Stations: leicester, kings lynn, cambridge, ...
Hotels: alpha-milton guest house, archway house, autumn house, ...
Hotel Types: hotel, guesthouse
Attractions: cambridge artworks, whale of a time, clare college, ...
Attraction Types: entertainment, mutliple sports, college, ...
Restaurants: ugly duckling, prezzo, caffe uno,
Cuisines: portuguese, vietnamese, italian, ..

When users mention locations or entity names that don't exactly match but are similar to these
options, try to match them to the closest valid option. For example:
- If user says 'city center', match it to 'centre'
- If user mentions 'kings cross', match it to 'london kings cross'
- If user says 'fitzwilliam', match it to 'fitzwilliam museum'

***Important Guidelines***
Decide the next step based on the user's question:
- Ask clarifying questions if the input is ambiguous.
- Use the provided tools appropriately. Do not make up tools.
- Always review the conversation history before responding.
- Be polite and professional in your responses.
- Respond concisely and clearly.
\end{verbatim} \\
\hline
\end{tabular}
\caption{System prompt for the Zero-shot MultiWOZ agent. An exhaustive list of entities were provided in the system prompt. For brevity, only a few are demonstrated.}
\label{tab:agent_system_prompt}
\end{table*}

\begin{table*}[t]
\small
\begin{tabular}{|p{0.95\textwidth}|}
\hline
\begin{verbatim}
***Task Description***
You are an expert in analyzing task-oriented dialogues. You are provided a conversation that contains
the user's input and the agent's response to the input. Your task is to analyze
the dialog acts present in the agent's response to a user input. A response can have multiple
dialog acts. For each dialog act, you must explain whether it is present or absent
in the agent's response.

***Dialog Act Types***
A dialog act represents the function of the agent's response. The possible dialog acts are:
- inform: Agent provides factual information or details relevant to the user query
- request: Agent asks user for specific missing information needed to fulfill their request
- select: Agent presents one or more options in response to the user's request and actively asks
user to choose between them
- recommend: Agent makes a specific suggestion with positive framing, often using phrases like
"I recommend", "I suggest", or "How about"
- nooffer: Agent explicitly states that no options exist matching user's criteria
- offerbook: Agent presents booking details and explicitly asks if user wants to proceed with
booking. Mostly occurs for train-related bookings
- book: Agent confirms a completed booking and usually provides certain details regarding the booking
- nobook: Agent states that a specific booking request cannot be completed
- greet: Agent responds to a user's indication that their needs have been met, typically 
accompanied by gratitude. The response can range from simple acknowledgments to well-wishes related
to the service provided or a simple goodbye.
- reqmore: Explicitly asks if user needs additional help or has other requests

***Few-Shot Examples***
Here are some examples of each dialog acts where the user input is followed by the agent's 
response and the dialog act is corresponding to the agent's response only:

**inform**

user: Could I have their phone number please and type of cuisine they serve?
agent: They serve international food and their phone number is 01223 812660.

user: Hello I would like to find a restaurant in the west part of the city that costs a moderate amount.
agent: Prezzo is a moderately priced italian restaurant in the west part of town.

**request**

user: Can you tell me what the travel time is and the price?
agent: Which train would you like that for?

user: It's important that I arrive no later than 23:30.
agent: Where will you be departing from?
...
...
...

******
Analyze the following conversation and list all dialog acts that apply. Multiple acts can be present
in a single response.

**Latest user input**:
{user_input}

**Latest agent response**:
{response}
\end{verbatim} \\
\hline
\end{tabular}
\caption{System prompt for WOZ dialog act classifier. 10 few-shot examples were used per act. For brevity, only a subset of the examples are demonstrated.}
\label{tab:dialog_act_prompt}
\end{table*}

\begin{table*}[t]
\small
\begin{tabular}{|p{0.95\textwidth}|}
\hline
\begin{verbatim}
***Task Description***
You are an expert in analyzing task-oriented dialogues. You are provided a conversation that contains
the current user input and an AI agent's response to it. Your task is to analyze the tools used by
the AI agent to generate the response. Multiple tools can be used in a single response. For each tool, 
you must explain if and why it has been used or not by the agent for the current response.

***AI Agent Description***
The AI agent whose response you are analyzing is a multi-domain customer support assistant for booking
various services in Cambridge, UK, including hotel bookings, restaurant reservations, train bookings, 
taxi bookings, and finding attractions. All the services and information it provides are specific
to Cambridge, UK and its surrounding areas.

***Available Tools***
The AI agent uses the following tools to generate responses:

1. **FindHotels**: The agent uses this tool to find hotels in Cambridge, UK. The agent provides
information about the hotel, such as its name, location, price, rating, etc. The agent generally
uses this tool to recommend one or more hotels to the user.
2. **BookHotel**: The agent uses this tool to book a hotel in Cambridge, UK. The agent may provide
information about the booking, such as the hotel name, check-in date, check-out date, number of guests, 
price, etc. The agent generally uses this tool to confirm that the booking was successful or 
not and provide details.
3. **FindRestaurants**: The agent uses this tool to find restaurants in Cambridge, UK. The agent 
provides information about the restaurant, such as its name, location, cuisine, price, rating, etc. 
The agent generally uses this tool to recommend one or more restaurants to the user.
4. **BookRestaurant**: The agent uses this tool to book a restaurant in Cambridge, UK. The agent 
provides information about the booking, such as the restaurant name, reservation date, reservation 
time, number of guests, price, etc. The agent generally uses this tool to confirm that the booking was
successful or not and provide details.
...
...
...

***Few-Shot Examples***
Here are some examples of each tool usage where the user input is followed by the agent's 
response and the agent uses that tool to generate its response:

**FindHotels**
user: Great! Yes, I'll also need to find a hotel with free parking and free wifi.
agent: I would recommend the ashley hotel.

**FindRestaurants**
user: Could I have their phone number please and type of cuisine they serve?
agent: They serve international food and their phone number is 01223 812660.
...
...
...

******
Analyze the following conversation and list all the tools used by the AI agent to generate the 
response. Multiple tools can be used in a single response. You must explain if and why each tool
has been used or not by the agent for the current response and provide a list of all tools
used by the agent for the current response based on your analysis.

**User input**:
{user_input}

**Agent response**:
{response}
\end{verbatim} \\
\hline
\end{tabular}
\caption{System prompt for the MultiWOZ tool classifier. 10 few-shot examples were used per tool. For brevity, only a subset of the examples are demonstrated.}
\label{tab:tool_use_prompt}
\end{table*}

\begin{table*}[t]
\small
\begin{tabular}{|p{0.95\textwidth}|}
\hline
\begin{verbatim}
***Task Description***: 
You are an expert task-oriented dialog system evaluator. Your task is to evaluate the agent response
and provide feedback on the quality of the response based on the conversation history and the
current user input.

***Evaluation Metrics***
1. Coherence (Flow and connection with context):
   Score 1: Completely disconnected from context, ignores previous conversation
   Score 2: Barely acknowledges context, major inconsistencies
   Score 3: Basic connection to context, some flow issues
   Score 4: Good connection to context, minor flow issues
   Score 5: Perfect flow and connection, fully integrates context

2. Specificity (Level of detail and precision):
   Score 1: Extremely vague, no concrete information
   Score 2: Minimal details, mostly general statements
   Score 3: Basic details, some specific information
   Score 4: Detailed information, most aspects covered
   Score 5: Comprehensive details, precise information about all aspects

3. Satisfaction (Likelihood of meeting user needs):
   Score 1: Completely fails to address user needs
   Score 2: Minimally addresses user needs, requires multiple follow-ups
   Score 3: Partially addresses needs, requires some clarification
   Score 4: Mostly addresses needs, minor clarification needed
   Score 5: Fully addresses all user needs, no clarification needed

4. Effectiveness (How well the response addresses the user input):
   Score 1: Completely misses the point of the input
   Score 2: Barely addresses the input, major gaps
   Score 3: Addresses main point but misses details
   Score 4: Addresses input well with minor omissions
   Score 5: Perfectly addresses all aspects of the input

***Evaluation Guidelines*** 
You have to evaluate the response using 4 metrics (coherence, specificity, satisfaction, 
effectiveness) on a scale of 1-5 for each metric. Follow these guidelines:
- Break down evaluation into specific aspects
- Consider both explicit and implicit requirements
- Provide concrete examples in reasoning
- Be consistent in scoring across evaluations
- Avoid position and length biases
- Avoid defaulting to the highest score; a lot of the times, responses will have some room for
improvement
- Be strict in your scoring. A score of 5 should only be given for truly exceptional responses that 
meet all criteria perfectly
- Focus on quality over quantity

******
Analyze the following conversation and provide a score for each of the evaluation metric. You must 
provide a reasoning to justify your score for each metric.

**Conversation History**
{context}

**Current User Input** 
{current_input}

Agent Response**:
{response}
\end{verbatim} \\
\hline
\end{tabular}
\caption{System prompt for the response quality evaluator.}
\label{tab:response_prompt}
\end{table*}

\end{document}